\documentclass[a4paper,fleqn]{cas-dc}

\usepackage[authoryear,longnamesfirst]{natbib}

\def\tsc#1{\csdef{#1}{\textsc{\lowercase{#1}}\xspace}}
\tsc{WGM}
\tsc{QE}
\tsc{EP}
\tsc{PMS}
\tsc{BEC}
\tsc{DE}

\begin{document}
\let\WriteBookmarks\relax
\def\floatpagepagefraction{1}
\def\textpagefraction{.001}

\shorttitle{Dynamic Graph Neural Network with Adaptive Features Selection for
RGB-D Based Indoor Scene Recognition}

\shortauthors{Q. Liu, R. Xiong, X. Chen M. Peng and Y. Yang}

\title [mode = title]{Dynamic Graph Neural Network with Adaptive Features Selection for RGB-D Based Indoor Scene Recognition}                      



%
\author[1]{Qiong Liu}
\author[1]{Ruofei Xiong}
\author[1]{Xingzhen Chen}
\author[1]{Muyao Peng}
\author[1]{You Yang}
\cormark[1]





\affiliation[1]{organization={School of Electronic Information and Communications,Huazhong University of Science and Technology},
    addressline={No. 1037, Luoyu Rd.}, 
    city={Wuhan},
    postcode={430074}, 
    country={P. R. China}}







\cortext[cor1]{Corresponding author: You Yang (Email: yangyou@hust.edu.cn)}



\begin{abstract}
Multi-modality of color and depth, i.e., RGB-D, is of great importance in recent research of indoor scene recognition. In this kind of data representation, depth map is able to describe the 3D structure of scenes and geometric relations among objects. Previous works showed that local features of both modalities are vital for promotion of recognition accuracy. However, the problem of adaptive selection and effective exploitation on these key local features remains open in this field. In this paper, a dynamic graph model is proposed with adaptive node selection mechanism to solve the above problem. In this model, a dynamic graph is built up to model the relations among objects and scene, and a method of adaptive node selection is proposed to take key local features from both modalities of RGB and depth for graph modeling. After that, these nodes are grouped by three different levels, representing near or far relations among objects. Moreover, the graph model is updated dynamically according to attention weights. Finally, the updated and optimized features of RGB and depth modalities are fused together for indoor scene recognition. Experiments are performed on public datasets including SUN RGB-D and NYU Depth v2. Extensive results demonstrate that our method has superior performance when comparing to state-of-the-arts methods, and show that the proposed method is able to exploit crucial local features from both modalities of RGB and depth.


\end{abstract}




\maketitle

\section{Introduction}

Scene recognition is among the most important and fundamental task in the community of computer vision~\cite{XIE2020107205,SINGH2020383,JING2020186,YANG2015INS,XIANG2021TIP}. Specifically, accurate scene recognition of contributes to the rapid development of applications such as indoor robots~\cite{liao2016understand,zhou2021borm,miao2021object,PEI2023TIMS}, scene parsing ~\cite{zhuo2017indoor,liu2019synthesis,fu2019adaptive,PEI2024TMM} and image retrieval~\cite{johnson2015image,sharif2019scene}. Specified on the problems, the research of indoor scenes struggle against the diversity of spatial layouts, complex lighting, and mutual occlusion among objects. With the availability of range sensors such as Kinect, researchers can easily have depth map accompany with RGB color image of the same scene for subsequent vision tasks~\cite{li2018cross,qi20173d,liu2020learning,george2021cross,PENG2025SR,PENG2025TIM}.  In this case, RGB-D indoor scene recognition has witnessed rapid development in recent years. However, capability varies on scene description among different modalities, and the challenge remains in how to properly treat different modalities in the task of scene recognition. In this context, exploring a discriminative scene representation to improve the recognition accuracy of RGB-D indoor scene recognition has become an urgent need for visual tasks.

It remains a challenge on how to select and treat features obtained from different modalities for RGB-D based scene recognition. Extant researches can be categorised by two parts, including global and local features. Most of previous works extract global features via convolutional neural network (CNN) for scene recognition~\cite{zhou2014learning,zhu2016discriminative,du2018depth,song2017combining,li2018df,du2019translate,HUANG2024SENSOR,XU2026RS}. For example, Li \textit{et al}.~\cite{li2018df} proposed to learn the correlative and distinctive features between two modalities simultaneously, and most of these features are global. Du \textit{et al}.~\cite{du2019translate} proposed to use high-quality RGB and depth images through the translate module, and simply combined the global features extracted from these two modalities for scene recognition. However, global features are good enough for scene classification in a general way, but not enough to represent the indoor scene with cluttered objects and complex spatial layouts.

Different from global features, other attempts have also been made on local features~\cite{wang2016modality,song2017depth}. Wang \textit{et al}.~\cite{wang2016modality} proposed to use object proposals extracted by fisher vector (FV)~\cite{sanchez2013image} coding to learn local features. Then, they combined the learned local features and the full-image based global features to get the final classification result. Song \textit{et al}.~\cite{song2017depth} extracted dense patches on the depth map as local features in a weakly-supervised manner and then fine-tuned the model with the full image for scene recognition. These methods all demonstrate the importance of local features for scene recognition. However, problem remains on how to select discriminative local features adaptively and exploit these features effectively for recognition. In particular, in multi-modality cases, the most challenging problem is on how to select helpful local features for scene recognition from numerous semantic information.

In recent years, many methods have focused on selecting discriminative local features. Song \textit{et al}.~\cite{song2017rgb,SONG2025CSVT} tried object detection on the scene first, and then extracted spatial location information to represent the scene. In addition, Song \textit{et al}.~\cite{song2019image} adopted co-occurring frequency and sequential representation of object location to describe relations among objects. Li \textit{et al}.~\cite{li2019mapnet} tried to extract a set of local semantic cues, and then used an intra-modality attentive pooling block to mine and pool discriminative semantic cues 
from each modality ~\cite{van2011segmentation}. In this method, cross-modality attentive pooling block was extended to learn different contributions across two modalities. Nevertheless, these two-stage approaches usually rely on the performance of object recognition and have higher computational complexity. Different from above methods, Xiong \textit{et al}.~\cite{xiong2019rgb} used non-local networks~\cite{wang2018non} as the spatial attention module to focus on key local features. Based on this, Xiong \textit{et al}.~\cite{xiong2020msn} proposed a framework with two branches to learn the multi-modality features. One branch extracted the local modality-consistent features of each modality by minimizing the similarity between the two modalities. The other branch learned the global modality-specific features between the two modalities by maximizing the correlation term. Xiong \textit{et al}.~\cite{xiong2021ask} used local features at both object-level and theme-level scales for scene recognition. However, the selected local features were only simply cascaded for scene recognition, and these approaches did not fully utilize local features. As a structured expression, graph neural network (GNN)~\cite{wu2020comprehensive,yan2018spatial,shi2020point} can flexibly represent the relations among local features, which is beneficial to connect local features with closer semantic or spatial distance. In addition, the aggregation and update operations of GNN can represent the information interaction among local features. Motivated by these capacities of GNN, Yuan \textit{et al}.~\cite{yuan2019acm} proposed to use a graph to model the relations among objects, but the different contributions between two modalities were not considered for scene recognition. Meanwhile, the connections among the nodes in the graph model were fixed which limited the performance of scene recognition. Therefore, extant works only focus on selecting local features without further considering the relations among local features, and thus there is a big room remained for the performance of RGB-D based scene recognition.

In this paper, we propose a dynamic graph model with an attention mechanism to handle the above problems. In the first step, we use an attention mechanism to adaptively select $k$ discriminative local features as nodes of the dynamic graph. In the construction of graph model, numerous interference information of indoor scenes is involved via global features. After that, nodes in graph are divided by three levels and then the graph model is dynamically updated according to attention weights by considering different contributions from various local features. It should be noted that node selection and graph update use two different attention mechanisms. The first one is used to extract discriminative local features to represent relations among local features from two modalities, and the second one is utilized for automatically learning the varying contributions of each modality when updating the node features. Finally, we fuse the optimized global and local features of both RGB and depth modalities, and then a linear classifier is used for scene recognition. Based on these modules, our method can adaptively extract discriminative local features, and experimental results show the effectiveness of our method. The contributions of our work can be summarized by two-fold.

\begin{enumerate}
	\item In order to effectively obtain discriminative local features, a dynamic graph is proposed to exploit local features effectively for scene recognition via two modalities, i.e., RGB and depth image. Specially, we propose an adaptive node selection method and adaptively compute a weighting parameter based on the global features, and the node with learned greater weights is used as the optimization node in the dynamic graph. 
	\item Different from most of the existing works that ignore the various contributions of the two modalities in different indoor scenes, the proposed model automatically learn different contributions of each modality when updating the node features in graph model.
\end{enumerate}

The rest of this paper is organized as follows. Sec. II describes the details of our method. Then, experimental results and discussions are presented in Sec. III. Finally, we conclude our work in Sec. IV.

\begin{figure*}[ht]
	\centering
	\includegraphics[scale=0.6]{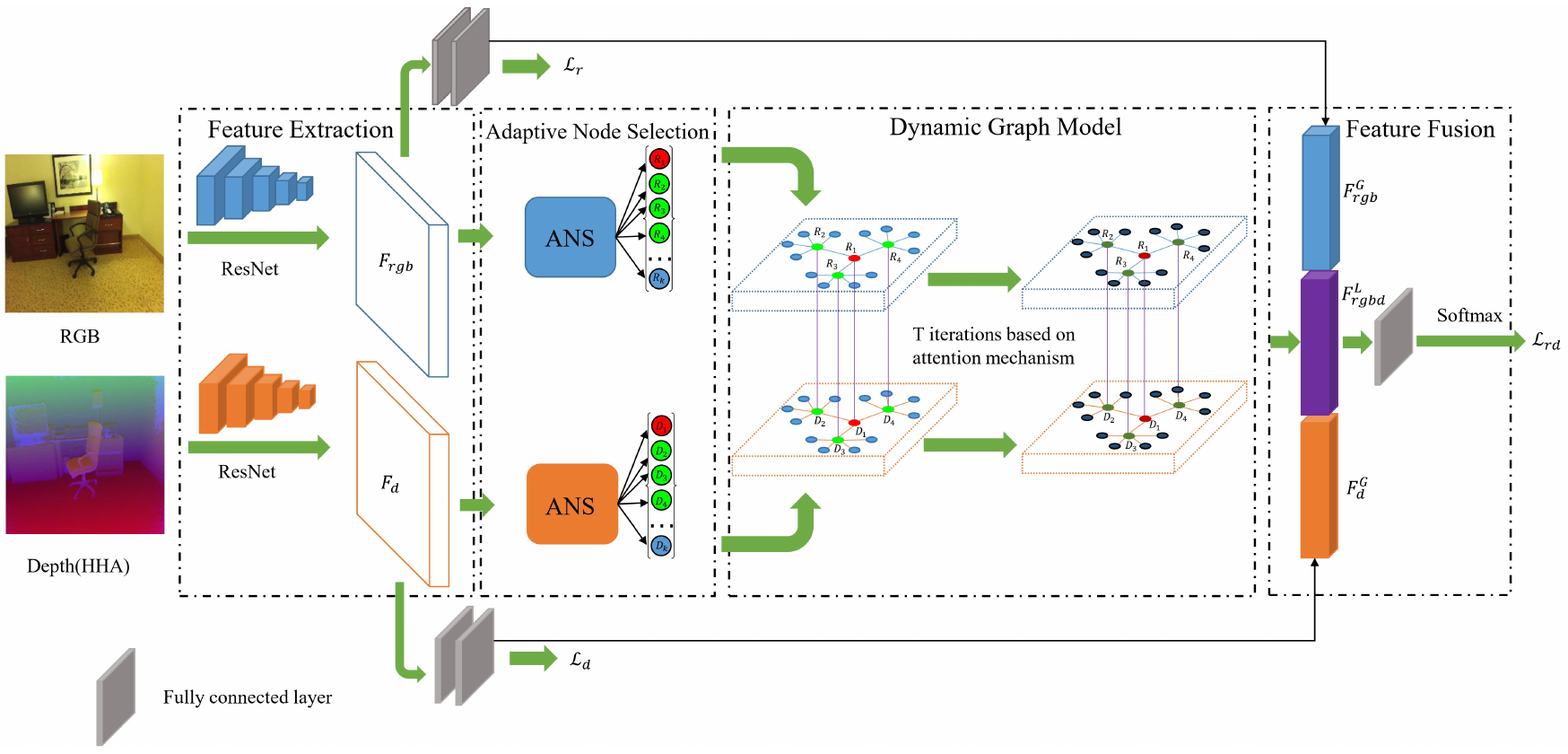}
	\caption{The framework of the proposed dynamic graph. First, RGB and depth images are input to two CNNs for feature extraction. Then, ANS module is used to select nodes to construct the multi-modality dynamic graph. In the last step, global and local features learned by the dynamic graph model are combined together for the final scene recognition.}
	\label{fig_framework}
\end{figure*}

\section{Dynamic Graph Model with Attention Mechanism}

\subsection{Framework of the Proposed Dynamic Graph Model}

An overview of the proposed framework is presented in Fig. \ref{fig_framework}. As depicted by this framework, the proposed model is organized by four modules, including Feature Extraction, Adaptive Node Selection, Dynamic Graph Model, and Feature Fusion. The most important thing in our work is the construction of graph model. For this purpose, both RGB and depth images are both input to two ResNets for global feature extraction in the feature extraction module. After that, an adaptive node selection (ANS) module is proposed to select discriminative local features as nodes for subsequent dynamic graph construction. The designed ANS is based on attention mechanism, which can effectively select the local features that beneficial for indoor scene recognition. The module of the dynamic graph model constructs connections between those selected nodes, and it is able to exploit relations among local features. In this module, the node features can be optimized and updated. After that, the global and local features after aggregation and update of the GNN are concatenated together in the module of feature fusion module for indoor scene recognition.

\subsection{Dynamic Graph Construction}
Specially, we construct a dynamic graph $G=(V, E)$ to represent relations among local features of the indoor scene. 
The node set $V$ are naturally grouped into two types: 2D appearance nodes $V_{rgb}$=\{$R_i|$ $i$=1,...,$k$\} and 3D geometry nodes $V_{d}$=\{$D_i|$ $i$=1,...,$k$\}, where $V$=$V_{rgb}$$\cup$ $V_{d}$ and $k$ is the number of selected nodes in each modality.
In addition, the edge set $E$ includes two subsets, i.e., intra-modality connections and inter-modality connections.
The first subset is denoted as $E_{rgb}$=\{$e_{ij}^{rgb}|j \in N(i)$\} and $E_d$=\{$e_{ij}^{d}|j \in N(i)$\}, where $E_{rgb}$($E_d$) represent RGB(depth) intra-modality connections and $j$ is a neighboring node of $i$. The second subset is denoted as  $E_{rgb-d}$=\{$e_{ij^{'}}^{rgbd}|j^{'} \in N(i)$\}, where $E_{rgb-d}$ represent that RGB and depth inter-modality connections and $i$($j^{'}$) belongs to RGB(depth) modality.

The selection of nodes and the connections of the dynamic graph are detailed below.

\subsubsection{Adaptive Node Selection} 
Global features can hardly fully represent the indoor scene image with cluttered objects and complex spatial layouts, and thus the contributions from local features in the scene should be considered. On the other hand, only discriminative local features should be involved in the dynamic graph modeling for recognition. In this case, contributions to indoor scene recognition should be evaluated on these local features.
Motivated by this, ANS is proposed to select local features as nodes of the dynamic graph for indoor scene recognition. This module is based on the attention mechanism, and
the computed attention map can represent the importance of local features for scene recognition. In addition, the ANS module helps to achieve end-to-end learning and has lower complexity compared to the two-stage methods.  

\begin{figure}[t]
	\centering
	\includegraphics[scale=0.3]{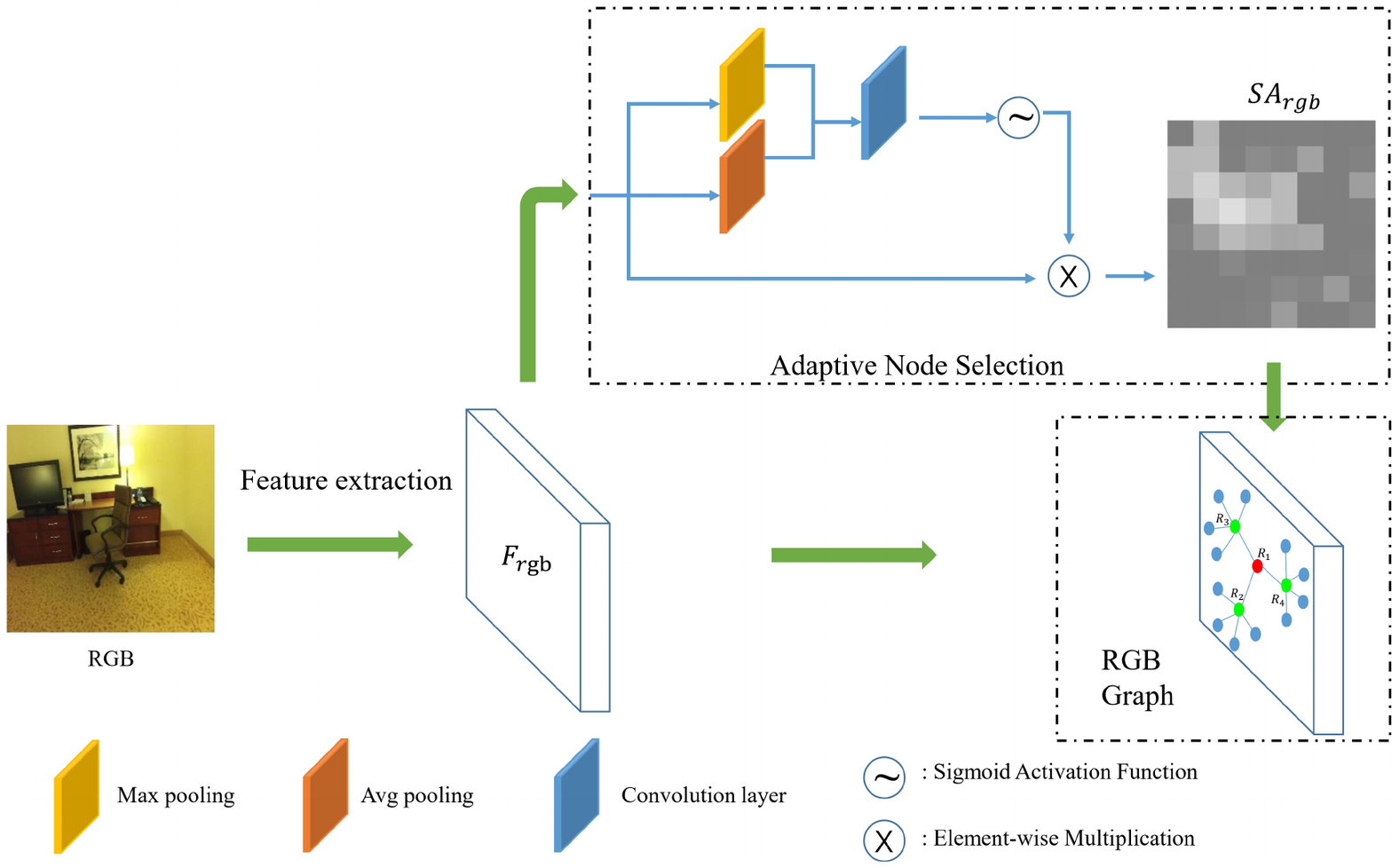}
	\caption{The selected key local features (i.e., nodes) for single modality (e.g., RGB) graph construction. The contribution of each node is different from the task of scene recognition, and it should be evaluated properly in graph modeling. In our dynamic graph model, the importance of each node is represented by its value in the computed attention map. By taking $k$ = 16 as an example, nodes are organized by 3 levels, including 1 main-central (i.e., in red), 3 sub-central (i.e., in green) and 12 leaf nodes (i.e., in blue), in the dynamic graph model. The three sub-central nodes \{$R_2$,$R_3$,$R_4$\} are connected to the main-central node $R_1$ and the rest of the leaf nodes are connected to the sub-central node by euclidean distance. Similar process is performed for depth dynamic graph construction. }
	\label{sig_mod}
\end{figure}

The attention module used in our dynamic graph model is adopted from CBAM~\cite{woo2018cbam}, which is a lightweight model. To compute the attention map, the average-pooling and max-pooling operations along the channel direction are used to learn the locations of discriminative local features effectively. Then the pooled features are concatenated and convolved by a standard convolution layer, producing the final attention map.
Specifically, $F_{rgb}$ and $F_{d}$ represent the last layer feature maps of two modalities. The tensor shapes of  $F_{rgb}$ and $F_{d}$ are $(B,C,H,W)$, where $B$ is the batch size, $C$ indicates the number of channels, and $H$,$W$ denote the height and width of the feature maps respectively. Then the attention map $SA_{rgb}$ of RGB modality is calculated by:
\begin{align}
	{SA_{rgb}} =\sigma(\theta(Concat(Avg({F_{rgb}}),Max({F_{rgb}}))))
\end{align}%
where $Avg$ and $Max$ indicate average-pooling and max-pooling operations, $Concat(\cdot , \cdot)$ denotes concatenation, $\theta$ is a convolution layer. $\sigma$ represents the activation function, which is applied to normalize the attention map. 
$SA_{d}$ can be computed from $F_{d}$ by the same process with $SA_{rgb}$.

Then the feature map $F_{rgb}^f$ enhanced by the attention mechanism is calculated as below:
\begin{align}
	F_{rgb}^f =F_{rgb}*{SA_{rgb}}+{F_{rgb}}
\end{align}%

The contribution of each node is different to scene recognition, and it should be evaluated properly and processed in dynamic graph modeling. In our proposed dynamic graph model, the importance of each node is represented by its value in the computed attention map. 
After calculating the attention map $SA_{rgb}$, we map the $k$ positions with the largest values in $SA_{rgb}$ to $F_{rgb}^f$ to obtain discriminative local features. Then, these local features are used as node attributes of the dynamic graph model.
Further, we divide nodes into three categories according to the importance level represented by the value of $SA_{rgb}$, including main-central node, sub-central nodes, and leaf nodes. 

Specifically, we first reshape the size of feature map $F_{rgb}^f$ from $(B,C,H,W)$ to $(B,H$$\times$$W,C)$. Then, by sorting the positions obtained by the attention map, we select the features from reshaped $F_{rgb}^f$ at the corresponding positions. Finally, we get a tensor of size $(B,k,C)$, denoted as $V_{rgb}$=\{$R_i|$ $i$=1,...,$k$\}. 
Meanwhile, we sort the $k$ nodes by $SA_{rgb}$ and take the first $m$ nodes as the most important level, the next $n$ nodes as the second level, and the rest as the third level. The typical values of $k$, $m$ and $n$ used in this paper are 16, 1, 3 respectively.

\begin{figure}[t]
	\centering
	\includegraphics[scale=0.3]{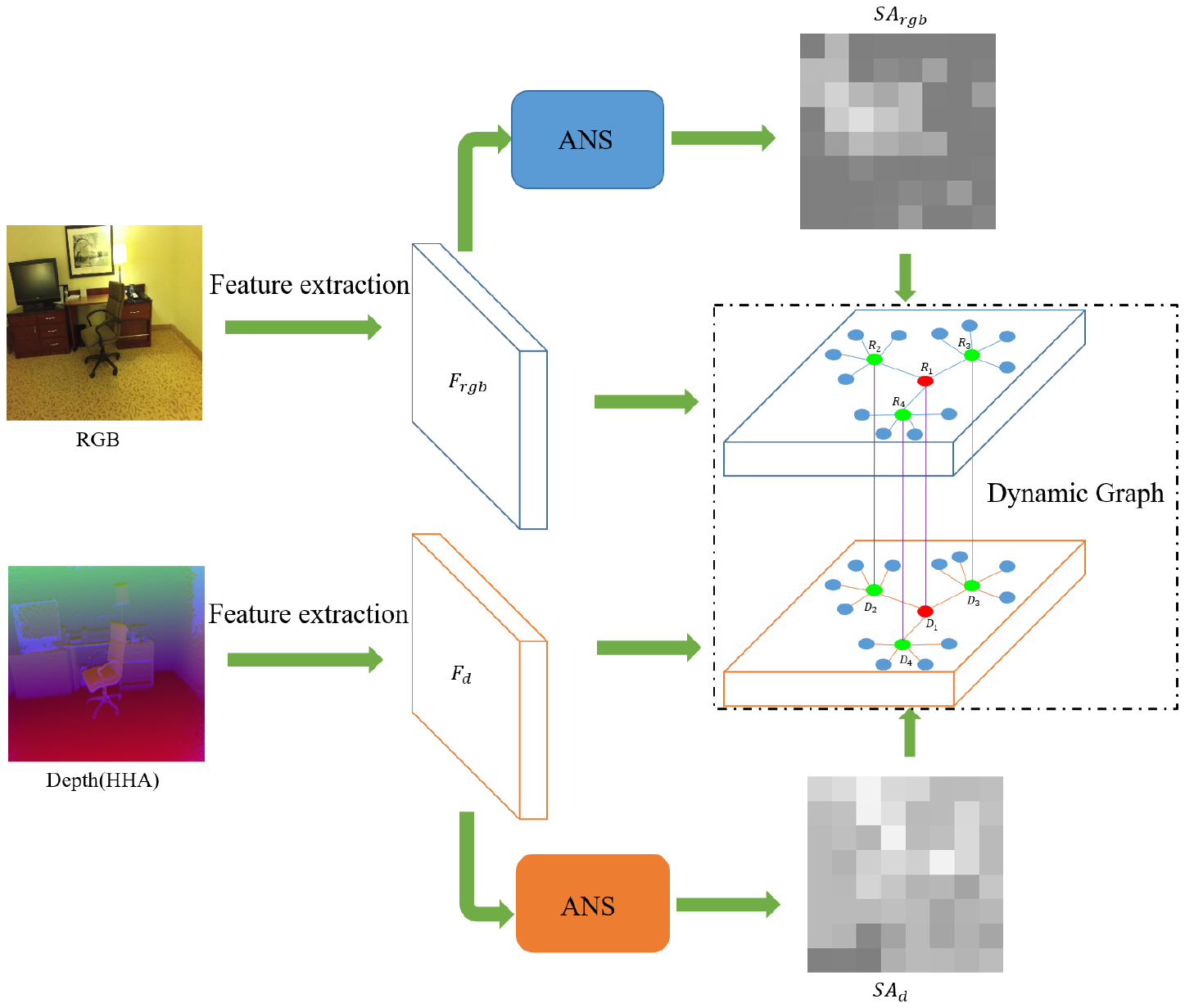}
	\caption{Connection between two modalities. Considering the semantic gap between the two modalities, a sparse connection is chosen which means main-central nodes, $R_1$ and $D_1$, sub-central nodes, \{$R_2$,$R_3$,$R_4$\} and \{$D_2$,$D_3$,$D_4$\} are connected separately.}
	\label{mul_mod}
\end{figure}
\subsubsection{Dynamic Graph Connection}  
As mentioned above, the dynamic graph model is organized into three levels. 
In this case, we construct sparse connections among nodes at three levels for the dynamic graph model.

More specifically, in order to construct intra-modality connections $E_{rgb}$=\{$e_{ij}^{rgb}|j \in N(i)$\}, the $n$ sub-central nodes are connected to the each main-central node. 
In addition, each leaf node is connected to the sub-central node with the closest Euclidean distance. By taking $k = 16$ as an example, the RGB modality connections is shown in Fig. \ref{sig_mod}.
 
In addition, in order to construct inter-modality connections $E_{rgb-d}$=\{$e_{ij^{'}}^{rgbd}|j^{'} \in N(i)$\}, we need to connect the corresponding nodes of the two modalities. Also considering the two modalities contains different features, a sparse connection is chosen which indicates that main-central nodes and sub-central nodes are connected separately. Specially, According to the computed attention map, the main center and sub-main center nodes of the two modalities, $R_i$ and $D_i$($i$=1,2,...,$m+n$), are connected in turn. By taking $k$ = 16 as an example, the inter-modality connections is shown in Fig. \ref{mul_mod}.

\subsection{Graph Aggregation and Update}

We use a dynamic graph to model relations among selected local features of the two modalities, and the two most essential aspects of GNN, including graph aggregation and update, are described below.

Specifically, we denote the feature of node $i$ as $h_{i}$ $\in$ $\mathcal{R}^{C}$($C$ is the number of channels in the above) in the dynamic graph and the features of its neighboring nodes \{ $h_{j}| j \in N(i)$\}. For the target node $i$, the features of neighboring nodes are first projected into the same feature space as node $i$ by a linear transformation $W$. Then these transformed features are aggregated with attention weights $\alpha$. The final information update is done by a non-linear activation function $\sigma$. ${l}$ denotes the ${l}-$th iteration. The whole process can be expressed as:
\begin{align}
	{h_i^{l+1}} =\sigma({h_i^{l}}+\sum_{j \in N(i)} \alpha_{ij}W{h_i^{l}})
\end{align}%
where the attention weight of target node $i$ and one of neighboring nodes $j$ can be calculated as
\begin{align}
	{e_{ij}} =w^T(Concat(W^{'}h_i,W^{'}h_j))
\end{align}%
\begin{align}
	{\alpha_{ij}}=softmax(e_{ij})
\end{align}%
where $w^T$ and $W^{'}$ respectively represent the weight matrices of different linear transformations, $softmax$ denotes softmax activation function.

Two attention modules are introduced in this paper, and it is necessary to emphasize that they have their specific functions. The first one is applied in ANS module which is used to extract discriminative local features in RGB(depth) modality as nodes of the dynamic graph. On the other hand, the next attention in the dynamic graph model is utilized for automatically learning the varying contributions of each modality when updating the node features.

In order to obtain the complementary information of the two modalities more effectively, the contributions of the local features of both two modalities are considered at the same time. The specific implementation is as follows:

\begin{align}
	{h_{{rgb}_i}^{l+1}} =\sigma({h_{{rgb}_i}^{l}}+\sum_{j \in N(i)} \alpha_{ij}W{h_{{rgb}_i}^{l}}+\sum_{j^{'} \in N(i)} \alpha_{ij^{'}}W{h_{d_j^{'}}^{l}})
\end{align}%

\begin{align}
	{h_{d_i}^{l+1}} =\sigma({h_{d_i}^{l}}+\sum_{j \in N(i)} \alpha_{ij}W{h_{d_i}^{l}}+\sum_{j^{'} \in N(i)} \alpha_{ij^{'}}W{h_{{rgb}_j^{'}}^{l}})
\end{align}%
where $h_{{rgb}_i}$, $h_{d_i}$ are the node features of the RGB and depth images respectively, $j$ and $j^{'}$ represent respectively the neighboring nodes of $i$ in same modality and different modality.

\subsection{Feature Fusion}
After the optimized features are obtained in dynamic graph, we use the RGB final feature map $F_{rgb}$ to get RGB global feature $F_{rgb}^{G}$ by a fully connected layer (FC).
After that, a cross entropy loss function is used to train the RGB global modality-specific feature, and the calculated loss is expressed as $\mathcal{L}_{rgb}^G$. 
The depth modality is processed in the same way to obtain depth global features $F_{d}^{G}$ from $F_{d}$ and corresponding loss $\mathcal{L}_{d}^G$.
Further, we concatenate the learned local features $F_{rgbd}^{L}$ by our dynamic graph model and global modality-specific features($F_{rgb}^{G}$ and $F_{d}^{G}$) as $F$ for final scene recognition.
\begin{align}
	F=Concat(F_{rgb}^{G},F_{d}^{G},F_{rgbd}^{L})
\end{align}%

The final recognition result can be predicted by an extra
cross entropy loss layer with loss $\mathcal{L}_{rgbd}^L$ after passing $F$
through another fully connected layer. The overall loss $\mathcal{L}$ can be formulated as:
\begin{align}
	\mathcal{L}=\mathcal{L}_{rgb}^G+\mathcal{L}_{d}^G+\mathcal{L}_{rgbd}^L
\end{align}%

In the test phase, we only use $F$ for final indoor scene recognition.

\section{Experiments and Discussions}
\subsection{Experiment Settings}
\subsubsection{Datasets}
Both SUN RGB-D~\cite{song2015sun} and NYU Depth v2~\cite{Silberman:ECCV12} are used in our experiments. The SUN RGB-D dataset~\cite{song2015sun}, including 10335 RGB and depth image pairs captured from a variety of different camera and depth sensors, is the largest publicly available dataset for RGB-D indoor scene recognition. We use 19 scene categories with 4845 images for training and 4659 images for testing to follow standard splits. On the other hand, NYU Depth v2~\cite{Silberman:ECCV12} contains 1449 images in total. The scene categories are grouped into ten, including 9 most common categories and the rest. To have a fare comparison with previous methods, we use 795 and 654 images for training and testing, respectively.

\subsubsection{Evaluation metric}
In order to evaluate the performance and then compare to state-of-the-arts, mean-class accuracy is used as the evaluation metric, which is calculated by averaging precision of all the categories as follows:
\begin{align}
	MeanAcc =\dfrac{1}{C} \sum_{c=1}^C\dfrac{correct_c}{Num_c}
\end{align}%
where $C$ is the total number of classes, $Num_{c}$ means the total number of class c, and $correct_{c}$ represent the number of correctly predicted samples of class c.

\subsubsection{Implementation Details}
\begin{table*}[t]
	\centering
	\caption{Performance comparison with the state-of-the-art
		methods on SUN RGB-D Dataset.}
	\begin{tabular}{c|c|c}
		\hline
		&Methods   & Mean-class Accuracy ($\%$) \\
		\hline
		\multirow{8}*{State-of-the-arts} 
		&Wang \textit{et al}.~\cite{wang2016modality} & 48.1$\%$  \\
		\cline{2-3}
		&Song \textit{et al}.~\cite{song2017rgb} &54.0$\%$ \\
		\cline{2-3}
		&Li \textit{et al}.~\cite{li2018df}         & 54.6$\%$      \\
		\cline{2-3}
		&Yuan \textit{et al}.~\cite{yuan2019acm}      & 55.1$\%$     \\
		\cline{2-3}
		&Li \textit{et al}.~\cite{li2019mapnet} & 56.2$\%$     \\
		\cline{2-3}
		&Du \textit{et al}.~\cite{du2019translate}     & 56.7$\%$      \\
		\cline{2-3}
		&Xiong \textit{et al}.~\cite{xiong2020msn}      & 56.2$\%$      \\
		\cline{2-3}
		&Xiong \textit{et al}.~\cite{xiong2021ask}  & 57.3$\%$      \\
		\hline
		{\bf~Proposed}&{\bf Dynamic Graph}			  & {\bf57.7$\%$}     \\
		\hline
	\end{tabular}
	
	\label{sun_rgbd}
\end{table*}

To generate the feature maps, ResNet101 was implemented with the Pytorch deep learning framework for both RGB and depth modalities. According to the general setup, the depth is encoded by \textit{Horizontal disparity, Height above ground, and Angle with gravity}  (HHA)~\cite{gupta2014learning} to make better use of depth information. The input image pairs are firstly resized to $256\times 256$, then random flip and random erasing are used for RGB and depth modality with a probability of 0.5. Normalization is also used in both train and test sets. The proposed dynamic graph model is trained for no more than 200 epochs, and the learning rate is initialized as 1e-5 and reduced by cosine learning rate decay. SGD optimizer is employed with a batch size of 4 to train the proposed framework.
\begin{figure}[t]
	\centering
	\includegraphics[scale=0.29]{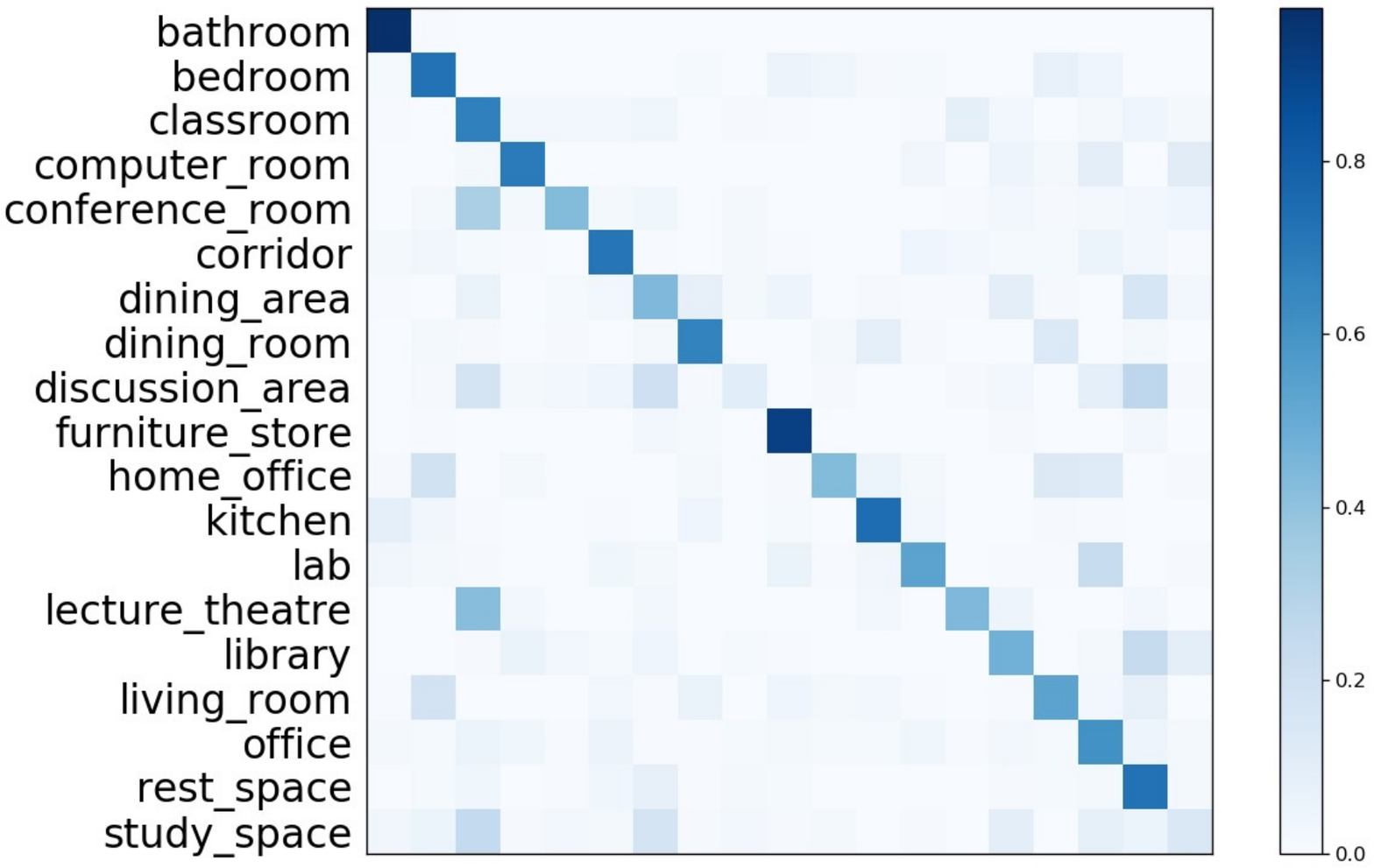}
	\caption{The classification confusion matrix of the proposed dynamic graph model on the SUN RGB-D Dataset. The vertical axis shows the ground-truth classes, the horizontal axis shows the predicted classes. The classes on the horizontal axis are in the same order as those on the vertical axis.}
	\label{sun_confuse}
\end{figure}
\begin{table*}[bp]
	\centering
	\caption{Performance comparison with the state-of-the-art
		methods on NYU Depth v2 Dataset}
	\begin{tabular}{c|c|c}
		\hline
		&Methods   & Mean-class Accuracy ($\%$) \\
		\hline		
		\multirow{8}*{State-of-the-arts} 
		&Wang \textit{et al}.~\cite{wang2016modality} & 63.9$\%$  \\
		\cline{2-3}
		&Song \textit{et al}.~\cite{song2017rgb} &66.9$\%$ \\
		\cline{2-3}
		&Li \textit{et al}.~\cite{li2018df}         & 65.4$\%$      \\
		\cline{2-3}
		&Yuan \textit{et al}.~\cite{yuan2019acm}      & 67.2$\%$     \\
		\cline{2-3}
		&Li \textit{et al}.~\cite{li2019mapnet} & 67.7$\%$     \\
		\cline{2-3}
		&Du \textit{et al}.~\cite{du2019translate}     & 69.2$\%$      \\
		\cline{2-3}
		&Xiong \textit{et al}.~\cite{xiong2020msn}      & 68.1$\%$      \\
		\cline{2-3}
		&Xiong \textit{et al}.~\cite{xiong2021ask}  & 69.3$\%$      \\
		\hline
		{\bf~Proposed}&{\bf Dynamic Graph}			  & {\bf70.4$\%$}     \\
		\hline
	\end{tabular}
	\label{nyu_rgbd}
\end{table*}

\subsection{Comparison to the State-of-the-art Methods} 
\subsubsection{Comparison Algorithms} 
Our method is compared to baseline and state-of-theart methods, including Wang \textit{et al}.~\cite{wang2016modality}, Song \textit{et al}.~\cite{song2017rgb}, Li \textit{et al}.~\cite{li2019mapnet}, Li \textit{et al}.~\cite{li2018df}, Du \textit{et al}. ~\cite{du2019translate}, Xiong \textit{et al}.~\cite{xiong2020msn}, Xiong \textit{et al}.~\cite{xiong2021ask} and Yuan \textit{et al}.~\cite{yuan2019acm}.
Specially, Li \textit{et al}.~\cite{li2018df} and  Du \textit{et al}. ~\cite{du2019translate} used only the global features in the scene for scene recognition. Wang \textit{et al}.~\cite{wang2016modality}, Song \textit{et al}.~\cite{song2017rgb} and Li \textit{et al}.~\cite{li2019mapnet} improved the performance of scene recognition based on object detection. Xiong \textit{et al}.~\cite{xiong2020msn} and Xiong \textit{et al}.~\cite{xiong2021ask} used attention mechanisms to select discriminative local features in the scene. In addition, Yuan \textit{et al}.~\cite{yuan2019acm} introduced GNN into the field of scene recognition for the first time, and used GNN to model the relations among objects.

\begin{figure}[t]
	\centering
	\includegraphics[scale=0.28]{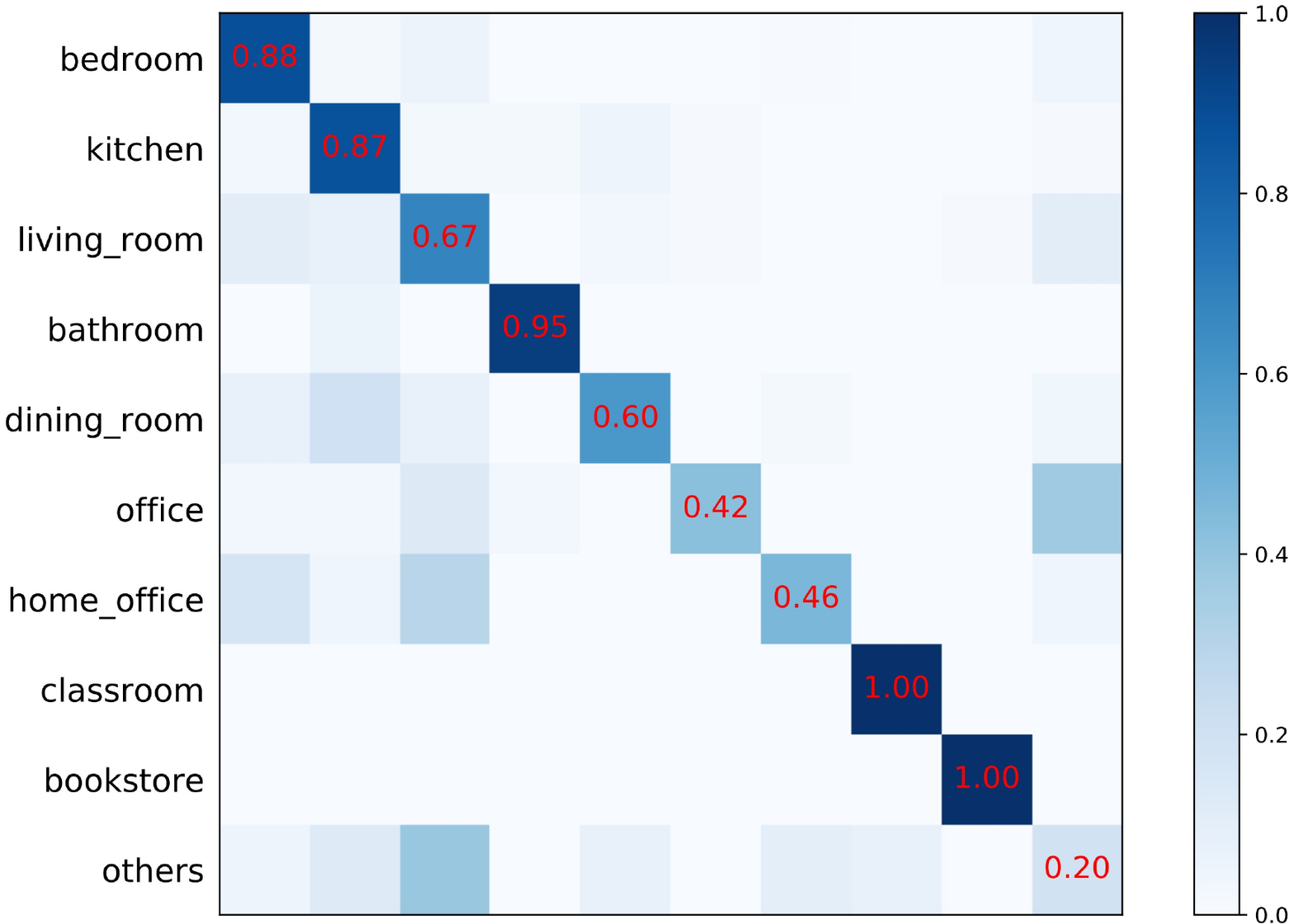}
	\caption{The classification confusion matrix of proposed dynamic graph model on the NYU Depth v2 Dataset. The vertical axis shows the ground-truth classes. The horizontal axis shows the predicted classes.}
	\label{nyu_confuse}
\end{figure}
\begin{table*}[bp]
	\centering
	\caption{Ablation study on SUN RGB-D Dataset.}
	\begin{tabular}{c|c|c}
		\hline
		Feature Types&Methods   & Mean-class Accuracy ($\%$) \\
		\hline
		\multirow{2}*{Single Modality without Dynamic Graph} 
		&Only RGB & 45.2$\%$  \\
		\cline{2-3}
		&Only depth &43.7$\%$ \\
		\hline
		\multirow{2}*{Single Modality with Dynamic Graph}
		&RGB Dynamic Graph         & 49.7$\%$      \\
		\cline{2-3}
		&depth Dynamic Graph      & 48.0$\%$     \\
		\hline
		\multirow{4}*{Multi-modality Fusion}
		&Mross-modality Simple Fusion     & 51.3$\%$      \\
		\cline{2-3}
		&{\bf Dynamic Graph}(without GNN)			  & {54.2$\%$}     \\
		\cline{2-3}
		&{\bf Dynamic Graph}(without Attention)			  & {55.5$\%$}     \\	
		\cline{2-3}
		&{\bf Dynamic Graph}($k$=16)		  & {\bf57.7$\%$}     \\		
		\hline
		
	\end{tabular}
	
	\label{sun_abla}
\end{table*}
\subsubsection{Comparison Results on SUN RGB-D Dataset} 
We first verify the effectiveness of our dynamic graph model on SUN RGB-D dataset. The comparison results are performed with state-of-the-art methods and given by Table \ref{sun_rgbd}. As depicted by Table \ref{sun_rgbd}, our proposed dynamic graph model has superior performance and achieves a mean-class accuracy of 57.7$\%$, which is better than all the state-of-the-art methods. These superior results show that ANS together with the optimization in dynamic graph model is able to figure out the discriminative local feature for subsequent indoor scene recognition. The proposed dynamic graph model is better than other graph and non-graph model based methods. The classification confusion matrix of our dynamic graph model on the SUN RGB-D Dataset is shown in Fig. \ref{sun_confuse}. Also can be depicted that, our method is able to recognize most of the scenes, while confused between \textit{discussion\_area} and \textit{rest\_space} because they are quite similar to each other.

\subsubsection{Comparison Results on NYU Depth v2 Dataset} 
In this section, we demonstrate the performance comparison with state-of-the-art methods on NYU Depth v2 dataset, as shown in Table \ref{nyu_rgbd}. In general, the experimental results on NYU Depth v2 dataset is similar to SUN RGB-D dataset. The proposed method outperforms all counterpart methods and achieves new state-of-the-art performance of 70.4$\%$ mean-class accuracy. Fig. \ref{nyu_confuse} shows the classification confusion matrix of proposed dynamic graph model on the NYU Depth v2 Dataset. We can see that the results of ``classroom", ``bookstore" and ``bathroom" are approaching to 100$\%$ accuracy. On the other hand, because of the small training images and the fact that "others" contains many challenge categories, our method has a lower performance in "others". 
\subsection{Ablation Study and Analysis}
To further illustrate the superiority of our proposed algorithm, we conduct the following ablation studies to explore the effect of different factors.
\subsubsection{Analysis of the Single- or Multi-Modality}
The Table \ref{sun_abla} and Table \ref{nyu_abla} show ablation study on SUN RGB-D dataset and NYU Depth v2 dataset. The first four rows of Table \ref{sun_abla} represent the accuracy of using a single modality, which includes the results with or without our proposed dynamic graph. ``Only RGB(depth)" indicates that only single modality global CNN features are extracted using the feature extraction network(ResNet101) for the final scene recognition. ``RGB(depth) Dynamic Graph" indicates the model using the dynamic graph on the only RGB(depth) images. The RGB image contains more information such as color and texture, while the depth image carries more information such as spatial geometry and location. From Table \ref{sun_abla}, we can observe that the ``Only RGB" model improves about 1.5$\%$ over ``Only depth" model and ``RGB Dynamic Graph" model improves about 1.7$\%$ over ``depth Dynamic Graph" model. We find that using a single RGB modality can achieve better accuracy than a single depth modality, which is also in line with the subjective perception of people.
\begin{table*}[t]
	\centering
	\caption{Ablation study on NYU Depth v2 Dataset.}
	\begin{tabular}{c|c|c}
		\hline
		Feature Types&Methods   & Mean-class Accuracy ($\%$) \\
		\hline
		\multirow{2}*{Single Modality without Dynamic Graph} 
		&Only RGB & 58.3$\%$  \\
		\cline{2-3}
		&Only depth &54.6$\%$ \\
		\hline
		\multirow{2}*{Single Modality with Dynamic Graph}
		&RGB Dynamic Graph         & 62.4$\%$      \\
		\cline{2-3}
		&depth Dynamic Graph      & 60.5$\%$     \\
		\hline
		\multirow{4}*{Multi-modality Fusion}
		&Multi-modality Simple Fusion     & 63.5$\%$      \\
		\cline{2-3}
		&{\bf Dynamic Graph}(without GNN)			  & {65.9$\%$}     \\	
		\cline{2-3}
		&{\bf Dynamic Graph}(without Attention)			  & {67.6$\%$}     \\	
		\cline{2-3}
		&{\bf Dynamic Graph}($k$=16)		  & {\bf70.4$\%$}     \\		
		\hline
		
	\end{tabular}
	\label{nyu_abla}
\end{table*}
\begin{table*}[t]
	\centering
	\caption{The effect of number of nodes on SUN RGB-D Dataset.}
	\begin{tabular}{c|c|c}
		\hline
		&The Number of Nodes & Mean-class Accuracy ($\%$) \\
		\hline
		\multirow{5}*{\bf Dynamic Graph} 
		&$k$=4		  & {53.6$\%$}     \\	
		\cline{2-3}
		&$k$=9			  & {55.5$\%$}     \\	
		\cline{2-3}	
		&$k$=16		  & {\bf57.7$\%$}     \\		
		\cline{2-3}
		&$k$=25			  & {56.1$\%$}     \\
		\cline{2-3}
		&$k$=36			  & {54.8$\%$}     \\
		\hline
		
	\end{tabular}
	
	\label{nodes}
\end{table*}
\subsubsection{Analysis of the Validity of Dynamic Graph Structure}
In our proposed dynamic graph model, the nodes are selected as vectors along the channel direction in the high-dimensional feature map. This choice helps to construct end-to-end learning tasks for RGB-D scene recognition. In addition, the contribution of each node to scene recognition is different, it needs to be treated separately. The importance of each node is determined by the attention map in ANS, in which a larger value means a greater contribution to scene recognition. 

In addition, we construct sparse connections among nodes for the dynamic graph model.

From the results,  we can observe that the model using dynamic graph on the only RGB(depth) images which is named ``RGB(depth) Dynamic Graph" improves about 4$\%$ over the general RGB(depth) model which is named ``Only RGB(depth)", from which we can also demonstrate the validity of the dynamic graph model. In addition, we use the model "Dynamic Graph(Without GNN)" with GNN removed, which indicates that local features are selected and then cascaded instead of using the dynamic graph model to construct the relations among local features. The mean-class accuracy decreases by 3.5$\%$, which also proves the effectiveness of the dynamic graph model.

\subsubsection{Analysis of the Multi-Modality Fusion}

In order to evaluate the effectiveness of multi-modality Fusion in our dynamic graph model, we conduct an ablation experiment by omitting the dynamic graph model from two modalities. Thus, Only the global features of the two modalities are used for fusion for the final scene recognition. We name this ablation model ``Multi-modality Simple Fusion" and the comparison results are present in Table \ref{sun_abla}.

From the results,  we can observe that merely employing single modality information can only achieve limited performance and multi-modality features are beneficial for scene recognition. Also, we find that multi-modality fusion will work better than mono-modality model, whether it is a simple fusion (``Multi-modality Simple Fusion") or a graph-based fusion (``Dynamic Graph"). Finally, we can observe 6.4$\%$(6.9$\%$) performance improvement of the dynamic graph-based model over the simple fusion model on SUN RGB-D dataset(NYU Depth v2 dataset). 
This is because the two modalities can provide complementary features. A single modality either RGB or Depth can lead to limited recognition performance. An effective fusion of the two modalities will contribute to the final scene recognition. It turns out that our proposed dynamic graph model can achieve this better.

\subsubsection{Analysis of Attention}

The proposed dynamic graph model is dynamic instead of a fixed connection. Specifically, our dynamic graph is initialized by a fixed connection, but it will change dynamically in the subsequent process. On the one hand, when performing node selection, the ANS model can dynamically select nodes to form a multi-modality dynamic graph. On the other hand, attention also automatically changes the weights of the connections among nodes when nodes are updated. We believe the setting of dynamic connection can be more adaptive to different types of scenes, and contribute to better recognition performance. 

To investigate the role of the attention mechanism in aggregation and update of node information, we set up an ablation test on the removal of attention (“Dynamic Graph(without attention)”) when $k$ = 16, and the results are given in Table II. As depicted by the results, the accuracy drops from 57.7$\%$ to 55.9$\%$ after attention module has been removed. This verifies our minds that dynamic connection and update of nodes are helpful to improve the performance of multi-modality based scene recognition.

\subsubsection{Analysis of the Number of Nodes}

To verify the impact of the selected number of nodes, we choose the number $k$ to be \{4, 9, 25, 36\} to compare the condition of $k$=16. It should be noted that the size of the attention map in this paper is $8\times 8$.  From Table \ref{nodes}, we can observe that $k$=4 performs worse than $k$=16 by 4.1$\%$ and $k$=9 performs worse than $k$=16 by 2.2$\%$, which is due to the fact that 4 or 9 nodes are not enough to represent a scene image. In addition, $k$=25 also performs worse than $k$=16 by 1.6$\%$ and $k$=36 performs worse than $k$=16 by 2.9$\%$, which also shows that it is not better to have more local objects to represent the scene, because it may bring unwanted noise and distracting information. In other words, we need to choose the right number of local features to represent the scene.

\section{Conclusion}
In this paper, we propose a dynamic graph model with attention mechanism for RGB-D indoor scene recognition.
Considering that indoor scenes usually contain numerous interference noises, we first use the adaptive node selection module to select $k$ discriminative local features as nodes of the dynamic graph. The module is based on the attention mechanism, and the importance of local features for scene recognition can be represented by the computed attention map.
To further utilize these features effectively, we divide nodes into three levels to describe the object corelation according to computed attention weights, and then our dynamic graph model is in sparse instead of full connection.
In addition, different from most of the existing RGB-D scene recognition works that different contributions of two modalities are not evaluated, the proposed dynamic graph model is able to automatically learn varying contributions of each modality when updating the node features. The experimental results show that we achieve the state-of-the-art results on both SUN RGB-D dataset and NYU Depth v2 dataset. Specifically, our dynamic graph model achieves 57.7$\%$ and 70.4$\%$ mean-class accuracy on the two standard datasets, respectively, outperforming all comparison algorithms. In future work, we will further consider how to better exploit the relations among local features in RGB-D indoor scene recognition and achieve a more meaningful fusion of the two modalities to overcome the semantic gap between the two modalities.

\printcredits

\bibliographystyle{cas-model2-names}

\bibliography{cas-refs}


\end{document}